# Becoming Suspicious Across Borders: Algorithmic Extraterritoriality and AI-Driven Financial Surveillance

**Georgios Pavlidis, Savvas A. Chatzichristofis and Eleni Gavriil**
Neapolis University Pafos, Cyprus

**Abstract**

Suspicion is an important, yet elusive concept in anti-money laundering and counter-terrorist financing (AML/CFT), which allows for intervention below the threshold of proof. In its traditional form, suspicion can be understood as a situated legal judgement by human actors within identifiable jurisdictions. It is argued that this understanding is no longer adequate. As artificial intelligence (AI) becomes an integral part of financial surveillance, suspicion is increasingly produced through data-driven processes. This transformation is epistemic, but also spatial. Since AI-driven financial surveillance operates through transnational data infrastructures, regulatory reach is less a matter of where conduct occurs than a question of whether such conduct becomes visible within data systems. This article develops the concept of algorithmic extraterritoriality, understood as a form of regulatory power mediated by data infrastructures rather than formal assertions of jurisdiction. Moreover, since individuals are increasingly constituted as datafied subjects of suspicion, they are rendered governable through dispersed and opaque processes of evaluation. This constitutes a challenge for accountability and contestability because suspicion becomes more difficult to locate, explain or contest.

## 1. Introduction: Becoming Suspicious Across Borders

Suspicion is a concept that occupies a rather uneasy space in law. It operates below the threshold of proof but above mere intuition, while it enables intervention without definitive knowledge.[1] In the context of anti-money laundering and counter-terrorist financing (AML/CFT), suspicion functions as a trigger, since it justifies monitoring, reporting and even exclusion from the financial system.[2] In its traditional form, suspicion can be understood as a situated legal judgement by human actors regarding specific transactions within identifiable jurisdictions.

We argue that this understanding of the term should be reconsidered. As artificial intelligence (AI) becomes an integral part of financial surveillance,[3] suspicion is no longer simply detected, but is increasingly a product of data-driven processes. Indeed, there is a marked shift from rule-based transaction monitoring towards deep-learning, graph-based and continuously updated detection models. In this context, financial institutions and financial supervisors can use transaction-monitoring systems, anomaly-detection tools and network analytics not only to identify pre-existing irregularities, but also to flag behaviours and ultimately individuals as suspicious.[4] In this sense, the AI tools analyse patterns, correlations and probabilistic inferences, which lead to the formulation of suspicion. Therefore, suspicion does not emerge from specific legal or factual triggers, but rather from computational infrastructures. It also shifts from an episodic assessment to a continuous condition. This suggests that AI does not simply improve the detection of pre-existing suspicious activity, but increasingly shapes the technical conditions under which suspicion is generated and acted upon.[5]

[1] In *Hussien v Chong Fook Kam* [1970] AC 942 (PC), Lord Devlin stated that 'suspicion in its ordinary meaning is a state of conjecture or surmise where proof is lacking … suspicion can take into account matters that could not be put in evidence at all'.
[2] De Koker, "Money Laundering Control."
[3] Soviany, "AI-Powered Surveillance."

At the same time, another development affects the spatial foundations of regulation. Legal authority is organised around territorial jurisdiction, even if it can extend extraterritorially through established doctrines such as effects, nationality or market access.[6] Nevertheless, AI-driven financial surveillance may operate through global data flows, which are not confined to territorial boundaries. In practice, financial institutions – especially in the context of financial groups – often aggregate and process data across jurisdictions, while their compliance systems rely on transnational information-sharing. Financial regulators and supervisors may also rely on cross-border data aggregation. This means regulatory reach can be exercised through the capacity to access, process and act upon data, and not only through formally asserted jurisdiction.

To capture this shift, this article advances the concept of ‘algorithmic extraterritoriality’. This concept refers to a regulatory reach that is not primarily grounded in legal doctrine, but instead in data infrastructures, as well as in the monitoring of individuals and transactions across borders. The concept of algorithmic extraterritoriality does not displace traditional forms of extraterritoriality,[7] but operates alongside them. In other words, what becomes visible through data becomes governable, regardless of where it is located.

These developments reconfigure how regulatory power is exercised in practice, and they also have implications for the human subject of financial regulation. Indeed, if suspicion is generated through probabilistic models and cross-border data aggregation, individuals can be constituted as subjects of suspicion independently of specific acts or locations. Moreover, since suspicion can result from data points dispersed across multiple jurisdictions, individuals do not simply become suspicious within a given legal order, but across borders. In such a diffuse and continuous space of data-driven governance, the subject’s experience of regulation shifts from specific encounters with legal authority to ongoing exposure to algorithmic classification.

We adopt a socio-legal and critical approach to examine these transformations, which take place at the intersection of law, technology and the human dimension. The focus is not on a purely doctrinal account of AML/CFT frameworks or the technical features of AI systems, but on the way data-driven infrastructures reshape the production of suspicion and the constitution of legal subjects. The central question guiding our analysis is: How does AI-driven financial surveillance transform the relationship between suspicion, jurisdiction and the human subject?

AML/CFT provides a particularly revealing setting for this transformation. Unlike many forms of predictive policing, financial surveillance is an integral part of routine private sector compliance, operates continuously across ordinary transactions and business relationships and relies heavily on cross-border information flows and group-wide compliance mechanisms.[8] It therefore combines preventive regulation, private gatekeeping and transnational data processing in a way that makes the relationship between suspicion and territorial jurisdiction especially visible.

To answer our key research question, Section 2 revisits the role of suspicion in law, while Section 3 examines how AI systems in AML/CFT produce suspicion through processes of classification and inference. Section 4 develops the concept of algorithmic extraterritoriality, arguing that regulatory reach is increasingly mediated by data infrastructures rather than territorial jurisdiction. Section 5 examines the implications of these developments for individuals, who can become datafied subjects of suspicion across borders. Section 6 examines the resulting challenges for accountability and legal governance, while also identifying tensions between surveillance practices and legal frameworks that are territorially bounded.

The article considers AI-driven financial surveillance as part of a broader transformation of legal space and knowledge, contributing to ongoing debates on law, technology and governance. It is argued that extraterritoriality should no longer be understood solely as a doctrinal question of jurisdiction, but also as a socio-technical condition, which depends on infrastructures of data and computation. This foregrounds an important problem in contemporary regulation: law continues to organise authority along territorial lines, but the production of suspicion and the exercise of regulatory power have become increasingly detached from such lines.

[4] Han, “AI for Anti-Money Laundering.”
[5] Fan, “Deep Learning Approaches”; Deprez, “Continual Graph Learning”; Mazumder, “Explainable and Fair AML Models.”
[6] Załucki, “Extraterritorial Jurisdiction.”
[7] Ryngaert, “Extraterritorial Enforcement Jurisdiction,” 539.
[8] Pavlidis, “The EU Anti-Money Laundering Directive.”

## 2. Suspicion in Law: From Situated Judgement to Infrastructural Condition

Suspicion has long functioned as a foundational concept in legal systems, although it can sometimes be elusive. Suspicion operates as a threshold device, which is lower than proof but sufficient to justify intervention. In the context of criminal law, suspicion enables investigative powers,[9] while in administrative law it can legitimise preventive action. In the context of financial regulation, suspicion triggers monitoring and reporting obligations,[10] while it can also lead to exclusion from financial services.[11] Despite the indeterminacy of suspicion, it is important to note that suspicion has traditionally remained anchored in judgement, which humans form through the interpretation of facts, contextualised within particular legal and institutional settings within identifiable jurisdictions.

Suspicion occupies a particularly central role in AML/CFT frameworks. Obliged entities – that is, financial institutions and other designated non-financial professions subject to AML/CFT obligations – are required to assess and monitor transactions and business relationships, as well as to report those deemed suspicious to competent authorities.[12] Such reporting is tied to identifiable events, which can be transactions, patterns of behaviour or deviations from expected norms within a given context. Formal rules are important in this process, but the formation of suspicion is not reducible to them. It relies on a combination of regulatory guidance, professional expertise and, most importantly, contextual awareness. Therefore, suspicion is neither fully subjective nor objective, but rather situated at the interaction between legal standards and human interpretation.

It can be argued that suspicion has two key characteristics: first, it is episodic, since it arises in response to triggers and it is assessed at specific moments in time; and second, it is situated, since it depends on knowledge of the context (identity of the customer, nature of the transaction, regulatory environment, etc.). These two characteristics are also present where AML/CFT obligations extend across borders, through mechanisms such as correspondent banking or international cooperation and sharing of information.

However, this model becomes problematic due to the growing scale, speed and complexity of financial transactions, which complicate assessments by human actors. Financial institutions must process large volumes of data across multiple jurisdictions, often in real time. In response, AML/CFT compliance had to incorporate automated systems, which are designed to assist, and in some cases to replace, human judgement.[13] Initially, AML/CFT compliance tools were designed to flag transactions that met predefined criteria. Such rule-based tools allowed for standardisation, but they largely maintained the underlying structure of suspicion as an event-based and context-dependent judgement. The integration of AI into AML/CFT systems was a step forward. Such AI systems do not operate through rule-based approaches, but instead through pattern recognition, statistical modelling and inferential analysis. In other words, they do not simply apply predefined rules to specific categories of risk, but identify correlations within large datasets. This may take place without clear *ex ante* definitions of what constitutes suspicious behaviour; therefore, suspicion does not result solely from observable deviations within the context of specific transactions. It can emerge from patterns detected across datasets, which may also span different institutions, jurisdictions and timeframes. This transformation is reinforced by the technical architecture of contemporary AML systems. Since money-laundering typologies evolve continuously, effective detection increasingly depends on models capable of adapting to new behavioural patterns without discarding previously learned structures. Suspicion therefore becomes continuous not only in regulatory effect, but also in computational maintenance.[14]

This has implications for the temporal and epistemic dimensions of suspicion. Temporally, suspicion becomes less episodic and more continuous. In other words, individuals and transactions are subject to ongoing evaluation, since risk scores and classifications are updated dynamically as new data become available. Epistemically, we move from interpretative judgement to probabilistic inference. AI systems do not ask whether a specific transaction appears suspicious in context, but they assess the likelihood that a specific pattern corresponds to known or inferred risk profiles. Therefore, the basis of suspicion becomes less transparent because complex models may not be readily interpretable even by those who deploy them.

---

[9] Skolnik, "Suspicious Distinction," 227.
[10] Simpson, "Role of Transaction Monitoring," 165.
[11] Malakoutikhah, "Financial Exclusion," 663.
[12] Financial Action Task Force, FATF Recommendations.
[13] Brewczyńska, "Combatting Financial Crime," 34.
[14] Fan, "Deep Learning Approaches "; Deprez, "Continual Graph Learning."

At the same time, it is more difficult to identify the locus of judgement, which becomes diffuse. Human oversight formally remains in place due to regulatory requirements,[15] but computational systems currently mediate the substantive process of identifying suspicion. The role of human actors is to review flagged transactions or validate outputs, but such role can be constrained by the scale of alerts and the opacity of underlying models. We argue that this can lead to deferred judgement, where algorithmic outputs shape and predetermine human decision-making.[16]

The effect of these developments is a gradual reconfiguration of suspicion from a situated legal judgement into an infrastructural condition. This means suspicion is not confined to judgements made by identifiable human actors, but becomes part of socio-technical systems that continuously process and evaluate data. We argue that these systems do not merely assist judgement; rather, they shape the conditions under which suspicion is generated and acted upon.

In its turn, this shift has implications for how suspicion is experienced and, more importantly, contested. In its traditional form, suspicion was – at least in principle – contestable because it could be traced to a particular decision, situated within a specific transactional context and a specific legal framework. In the context of AI-driven financial surveillance, suspicion becomes more difficult to locate and challenge because it is generated through dispersed and continuous data processing. Indeed, suspicion may not correspond to any single act or event, but may attach to patterns inferred across multiple data points and contexts. To understand this shift, it is worth examining how AI systems operate and how they produce the patterns and inferences that underpin contemporary financial surveillance.

## 3. AI, Financial Surveillance and Suspicion

Within the AML/CFT framework, obliged entities are required to conduct ongoing customer due diligence and transaction monitoring, and to report suspicious transactions or activities to the competent financial intelligence unit. Suspicion performs an operational regulatory function: it connects private monitoring by obliged entities with potential investigation and enforcement by public authorities.[17] Although the precise reporting threshold varies across legal systems, the FATF Recommendations provide the international framework around which these obligations are organised.[18]

Due to the increasing deployment of AI in AML/CFT systems, suspicion may be produced actively through computational processes, which changes the epistemic foundations upon which suspicion rests. AML/CFT systems rely increasingly on AI-driven techniques, such as machine learning models, anomaly-detection systems and network analytics, to process vast quantities of transactional and behavioural data. The objective is to identify patterns that may indicate illicit financial activity.[19] AI systems operate through inductive reasoning and infer patterns from data, but are not tied to predefined thresholds or known typologies as in the earlier rule-based approaches. In this process, there is often no stable definition of what constitutes suspicious behaviour in advance. At the technical level, the current state of the art increasingly relies on relational models, graph-based learning and complex network approaches, because suspiciousness is often embedded in transactional structures rather than isolated events. Indeed, recent studies use graph neural networks, topology-aware graph models, community detection and hybrid deep-learning pipelines to identify hidden relational patterns that conventional rule-based monitoring cannot capture effectively.[20]

In rule-based systems, a transaction is flagged because it matches a known risk indicator. Therefore, suspicion is anchored in recognition. In AI-driven systems, behaviours are rendered suspicious because they resemble patterns identified within data, although those patterns may be contingent, evolving or partially understood. Therefore, suspicion emerges from correlation and statistical association rather than from contextual interpretations. For its part, the object of suspicion expands because it no longer attaches solely to specific transactions or acts, but instead to patterns, relationships and networks. A single transaction that may appear unremarkable in isolation could become suspicious when considered within a broad constellation of data points

[15] Enqvist, “Human Oversight,” 508.
[16] Pomerol, “AI and Human Decision Making.”
[17] Pavlidis, “The EU Anti-Money Laundering Directive and Regulation.”
[18] The Financial Action Task Force (FATF) is the principal international standard-setting body for combating money laundering, terrorist financing and proliferation financing. Its Recommendations are soft law instruments that provide the global framework for AML/CFT regulation, while its mutual evaluation and monitoring processes assess jurisdictions’ technical compliance and the effectiveness of their systems, exerting significant influence on domestic and regional regulatory reform; see Financial Action Task Force, FATF Recommendations.
[19] Rahman, “Machine Learning and Network Analysis.”
[20] Oliveira, “Complex Networks-Based Anomaly Detection”; He, “Explainable Graph Neural Network Framework”; Fan, “Deep Learning Approaches.”

(geographic, behavioural or inferred associations with other persons). Suspicion, in this sense, is produced through the aggregation and analysis of data across time and space.

Regarding the element of temporality, suspicion is continuously recalibrated by AI-driven tools rather than being triggered by specific events. This means individuals and transactions are subject to ongoing evaluation and risk scores are updated in a dynamic manner as new data enter the system. Since models and datasets evolve constantly, the distinction between 'normal' and 'suspicious' becomes fluid and contingent on this evolution. Given this evolution, suspicion is an ever-present possibility, rather than an exceptional state, within the routine operation of financial infrastructures. Such temporal fluidity is particularly important in newer AML systems designed to detect emerging schemes rather than only previously labelled behaviours. Recent hybrid and unsupervised approaches explicitly aim to identify novel laundering patterns under conditions of class imbalance, partial visibility and rapidly shifting behavioural signatures.[21]

Moreover, there are legitimate concerns about the opacity surrounding the production of suspicion. AI systems – particularly those based on complex machine learning techniques – often generate outputs that are difficult to interpret or explain, functioning as 'black boxes'.[22] Typically, explainability tools do not provide full accounts of how a particular inference was reached, instead arriving at approximations.[23] Such opacity may complicate both internal decision-making and external accountability.[24] There is a risk that financial institutions rely on system outputs without fully understanding the system's underlying logic. For their part, individuals may be affected by these decisions without the possibility of contesting them effectively.[25] In this context, it is important to treat explainability, auditability and fairness as operational requirements rather than optional design preferences. This is especially important where institutions are expected to justify alerts, calibrate thresholds, document model behaviour and show that suspiciousness is not inferred through hidden discriminatory proxies. In this sense, explainability can be viewed not simply as an epistemic virtue, but as part of the institutional legitimacy of suspicion.[26]

The regulatory environment in which AML/CFT systems operate may reinforce the opacity of AI-driven suspicion. Confidentiality obligations, especially prohibitions on 'tipping off', limit the extent to which the basis of suspicion can be disclosed to affected individuals.[27] This creates a problem of structural asymmetry, because suspicion is generated within complex data-driven systems while remaining largely inaccessible to affected persons. It could be argued that this is a form of asymmetric visibility, in which individuals become increasingly transparent to institutions while the processes generating suspicion remain opaque.

It is also important to note that AI-driven systems are not neutral tools. Their outputs are shaped by the data on which they are trained, the design choices in their architecture and the regulatory priorities in each jurisdiction.[28] For example, algorithmic processing may reproduce and amplify biases resulting from training data.[29] This risk is present in the context of financial surveillance, since certain individuals, groups or regions may be disproportionately rendered suspicious due to biases reflecting historical inequalities, institutional practices or geopolitical issues. This concern is not merely theoretical; recent work in AML has shown that predictive gains may be sought through forms of customer profiling that rely on culture-linked or jurisdiction-linked variables, raising serious questions about proportionality, fairness and the legitimacy of embedding such proxies into systems that shape suspicion.[30]

There are also implications regarding the boundary between detection and prevention, which seems more blurred. Indeed, AI systems have the potential to enable pre-emptive intervention, often before any unlawful activity has occurred, since they identify patterns associated with potential risk.[31] In this configuration, suspicion becomes forward-looking since it is not formulated through the assessment of past behaviour, but the anticipation of risk. This enhances the preventive capacity of AML/CFT frameworks, but it raises questions about the threshold for intervention, while individuals may be subject to regulatory measures based on probabilistic assessments rather than concrete actions.

---

[21] Kungu, "Hybrid Deep Learning "; Fan, "Deep Learning Approaches."
[22] Von Eschenbach, "Transparency and the Black Box Problem," 1607.
[23] Dwivedi, "Explainable AI," 5.
[24] Chesterman, "Through a Glass, Darkly," 271.
[25] Novazi, "Automated Decision-Making," 149.
[26] Mazumder, "Explainable and Fair AML Models"; Hanson, "Law-Machine Learning Translation Problem."
[27] Dabor, "Examination of the UK SAR Regime," 107.
[28] Hacker, "Legal Framework," 259.
[29] Ntoutsi, "Bias in Data-Driven AI Systems," 3.
[30] Goodell, "'Profiling' and AML."
[31] Tumin, "Age of Pre-Emptive Power."

Regarding the spatial reach of regulation, suspicion increasingly enables regulatory influences that extend beyond traditional jurisdictional boundaries because it is generated through cross-border data flows. Suspicion is formulated through infrastructures that operate across institutions, jurisdictions and temporal scales, which renders it difficult to localise. Territorial presence still affects the capacity to act upon suspicion, but such capacity becomes less dependent on territory and more on access to data and to infrastructures. The next section examines this shift from territorial to infrastructural forms of regulatory power in more detail.

## 4. Regulatory Reach and the Concept of Algorithmic Extraterritoriality

The transformation of suspicion, which was examined in the previous sections, is not only a question of epistemology or technology; it is a development that reconfigures the spatial dimension of legal authority. The mechanisms through which suspicion is produced transcend territorial boundaries, as the process becomes data-driven and continuous. This calls into question the adequacy of understandings of extraterritoriality in its traditional legal sense. Indeed, extraterritoriality refers to the extension of regulatory authority beyond the territorial borders of a specific state.[32] It is typically based on principles such as the effects doctrine, nationality, protective jurisdiction or access to domestic markets. In this conventional legal sense, extraterritoriality remains tied to a jurisdictional connecting factor through which a state justifies prescribing, adjudicating or enforcing rules in relation to conduct beyond its territory. Its legitimacy and limits are therefore ordinarily assessed by asking whether the exercise of authority can be attributed to a state and justified by an accepted basis of jurisdiction.[33]

Traditional debates on extraterritoriality frequently foreground interstate concerns, particularly the risk that one state's regulatory action encroaches upon another state's sovereignty. The concern developed here is different but complementary: the analysis focuses primarily on what infrastructurally mediated regulatory reach means for individuals whose data and conduct become subject to transnational processes of classification and suspicion, possibly without any accountability mechanism.[34]

Extraterritoriality is a defining feature of AML/CFT frameworks, due to the complex transnational nature of the phenomenon of money laundering. On the one hand, financial institutions and other obliged entities operating internationally must comply with multiple regulatory regimes. On the other hand, supervisory authorities must rely on cooperation mechanisms and international standards, primarily those developed by the Financial Action Task Force (FATF),[35] to ensure a degree of global alignment.

Nevertheless, existing forms of extraterritoriality in the AML/CFT framework presuppose identifiable jurisdictions and clearly bounded legal orders. In other words, when regulatory influence extends across borders, it moves through recognisable legal channels: legislation, enforcement actions or conditions for market participation. Thus, it remains territorially anchored. For its part, AI-driven financial surveillance does not operate within a single territorial space, but aggregates and analyses data across jurisdictions. It functions through distributed infrastructures that connect financial institutions, data repositories and regulatory expectations. This means its capacity to generate suspicion is no longer dependent on territorial presence, but on data, computational resources and networked systems.

This article considers this transformation and advances the concept of algorithmic extraterritoriality, which is defined as a form of regulatory reach exercised through the capacity to access and process data across borders. While traditional extraterritoriality is grounded in legal doctrine, algorithmic extraterritoriality is mediated by socio-technical infrastructures. It operates through the practical ability to render activities visible within data-driven systems.

The actors exercising this form of regulatory reach are not limited to states or public authorities. In AML/CFT, financial institutions act as regulatory gatekeepers by deploying monitoring systems and generating or operationalising suspicion, while technology providers may shape the infrastructures through which this occurs. Public authorities, in turn, establish the legal obligations and supervisory expectations within which these systems operate. Algorithmic extraterritoriality should therefore be understood as a distributed form of regulatory power involving both public and private actors.

Algorithmic extraterritoriality manifests in several ways. First, financial institutions may centralise data processing across jurisdictions, often using global compliance platforms. This means transactions in one country may be analysed in another, while models are trained on datasets that incorporate information from multiple regions. Suspicion can therefore be generated through processes that are inherently transnational, although a transaction or activity may be localised.

---

32 Kamminga, "Extraterritoriality."
33 Kamminga, "Extraterritoriality"; Załucki, "Extraterritorial Jurisdiction"; Ryngaert, "Extraterritorial Enforcement Jurisdiction."
34 See Ireland-Piper, "Accountability in Extraterritoriality", examining extraterritorial jurisdiction through the lens of jurisdictional restraint, accountability and individual rights, and highlighting the legal uncertainty generated by competing assertions of jurisdiction.
35 Murrar, "Role of FATF," 77.

Second, AI-driven financial surveillance allows regulatory expectations to travel beyond traditional jurisdictional boundaries. More specifically, financial institutions and other obliged entities in jurisdictions with stringent AML/CFT regulations may apply uniform compliance standards across their global operations. For example, the EU AML/CFT regime imposes on obliged entities the obligation to implement group-wide AML/CFT policies, which in practice favour the use of global compliance platforms.[36] Similar developments diffuse regulatory norms beyond formal jurisdictions. In other words, entities operating in less regulated environments may be subject to stringent monitoring and reporting, not because there is a direct legal obligation but because they are integrated into global financial networks. This infrastructural account of regulatory reach is reinforced by recent EU reforms. The recent EU AML/CFT package strengthens a more integrated supervisory architecture through a single rule book, the creation of the Authority for Anti-Money Laundering and Countering the Financing of Terrorism (AMLA) and an expanded logic of cross-border coordination. In practical terms, this increases the importance of shared compliance infrastructures, standardised monitoring practices and group-wide governance arrangements as vehicles through which regulatory influence travels across jurisdictions.[37]

Third, risk scores, alerts and, more broadly, the outcomes of AI-driven analysis may circulate across institutional and jurisdictional boundaries. This happens through information-sharing mechanisms, both in the context of intergovernmental cooperation and on an informal basis, such as through intra-group data transfers.[38] The result is a form of cross-jurisdictional regulatory interdependence, in which an algorithmic assessment generated in one institutional or legal context shapes decisions taken in another.[39]

Consider, for example, a customer whose otherwise lawful transaction in jurisdiction A is processed through a banking group’s compliance infrastructure in jurisdiction B. A model trained partly on cross-border data may assign heightened risk because of the customer’s transactional network or geographic associations; that classification may then circulate within the group and contribute to enhanced monitoring, delayed transactions or de-risking elsewhere. The regulatory effect experienced by the individual thus results not from a single extraterritorial legal act, but from the transnational operation of interconnected data and compliance infrastructures.

For these reasons, it can be argued that, in addition to territorial jurisdiction, data visibility must be considered as a determinant of regulatory reach. What can be captured and processed within AI-driven financial surveillance becomes subject to governance. Conversely, what remains outside the data infrastructures may escape regulatory scrutiny. Thus, contours of data flows rather than the boundaries of states can determine regulatory power. This has important implications for the concept of extraterritoriality. Indeed, algorithmic extraterritoriality is not a simple extension of sovereignty, but reflects a more diffuse and less visible form of power. Instead of coercion or formal enforcement, the key elements are infrastructural integration and exposure to transnational surveillance and classification mechanisms.

Algorithmic extraterritoriality complicates questions of accountability. Since the formulation of suspicion may span multiple jurisdictions, it may be difficult to identify the locus of decision-making. In other words, responsibility is also dispersed across institutions and technologies. Traditional legal tools, in particular jurisdictional rules[40] or conflict-of-laws principles,[41] cannot easily be adapted to address these forms of distributed responsibility. However, it is important to note that algorithmic extraterritoriality does not replace traditional forms of extraterritoriality, which continue to play a significant role in global financial regulation.[42] What changes is the fact that data-driven systems reshape how regulatory power is exercised. The new relationship between governance and infrastructure thus supplements, and in some cases displaces, the traditional relationship between law and territory. For their part, individuals seem to be drawn into a transnational space of surveillance in which suspicion is continuously generated and circulated.

[36] See, for example, Regulation (EU) 2024/1624, arts 16ff.
[37] Regulation (EU) 2024/1624, arts 16ff.; Regulation (EU) 2024/1620; Brewczyńska, “Combatting Financial Crime.”
[38] Massa, “Information Flows,” 288.
[39] Regulation (EU) 2024/1624; Horobets, “Artificial Intelligence Technology.”
[40] Krisch, “Jurisdiction Unbound,” 582.
[41] Niboyet, “Territoriality and Conflict of Laws,” 582.
[42] Lehmann, “Legal Fragmentation.”

In this respect, algorithmic extraterritoriality is not only a question of where authority is asserted, but also of how legal norms are translated into the model architectures of AI, as well as the risk indicators, thresholds and validation criteria. The infrastructures through which suspicion travels are therefore also infrastructures through which law is operationalised.[43]

The preceding analysis suggests that algorithmic extraterritoriality involves interconnected changes in the locus, temporal logic, epistemic basis and object of suspicion, as well as in the position of the regulated subject and the structure of accountability. Table 1 synthesises these dimensions by contrasting the ideal-typical characteristics of traditional AML/CFT suspicion with those emerging in AI-driven financial surveillance.

**Table 1. From situated suspicion to algorithmic extraterritoriality: the transformation of suspicion in AI-driven financial surveillance**

| Dimension | Traditional AML/CFT suspicion | AI-driven financial suspicion | Legal/governance implication |
|---|---|---|---|
| **Locus of judgement** | Human actors assess concrete transactions in context. | Computational systems mediate and pre-structure suspicion through models, alerts and scores. | Responsibility becomes diffuse and harder to locate. |
| **Temporal logic** | Episodic, triggered by specific events or deviations. | Continuous, dynamically updated as data and models evolve. | Suspicion becomes an ongoing condition rather than an exceptional assessment. |
| **Epistemic basis** | Contextual interpretation and professional judgement. | Correlation, probabilistic inference and pattern recognition. | Grounds of suspicion become less transparent and less contestable. |
| **Object of suspicion** | Specific acts, transactions or identifiable conduct. | Patterns, relationships, networks and inferred associations. | Individuals may be rendered suspicious through relational positioning rather than direct action. |
| **Spatial logic** | Anchored in identifiable jurisdictions and legal orders. | Produced through cross-border data infrastructures and transnational compliance systems. | Regulatory reach becomes increasingly infrastructural rather than purely territorial. |
| **Subject position** | Legal subject situated within a recognisable decision context. | Datafied subject constituted through profiles, scores and fragmented data points. | Individuals are increasingly governed through opaque classificatory processes. |
| **Mode of intervention** | Reactive monitoring and reporting. | Predictive, preventive and pre-emptive identification of potential risk. | Thresholds for intervention shift closer to anticipated rather than completed conduct. |
| **Accountability structure** | Traceable decision-maker and clearer site of challenge. | Multiple actors, outsourced models, dispersed infrastructures and indirect scoring effects. | Review, explanation and redress become fragmented and weaker. |

As Table 1 makes clear, this shift affects not only the production of suspicion, but also the constitution of the subject and the distribution of accountability. The next section examines how this affects the constitution of the legal subject and the concept of the datafied subject of suspicion across borders.

[43] Hanson, "Law-Machine Learning."

## 5. The Datafied Subject: Becoming Suspicious Across Borders

In traditional legal frameworks, the subject of suspicion can be identified and situated within a given context. Suspicion attaches to a person in relation to specific conduct, within a defined legal environment. Even where the assessments are imperfect or contestable, they attach to a recognisable relationship between the individual, the alleged behaviour and the authority exercising judgement. Therefore, the individual can, in principle, locate the source of suspicion and engage with it through legal mechanisms.

By contrast, AI-driven financial surveillance reshapes the subject of regulation. Individuals are increasingly treated as datafied subjects,[44] which are evaluated and governed through fragments of information. Within AI-driven AML/CFT systems, processes of data aggregation and algorithmic inference define the subject, representing it not as a coherent legal actor, but as collections of data points. These data points, which may include transaction histories, geographic indicators, behavioural patterns, network connections and so on, are processed across systems and jurisdictions, generating profiles. The outcome may bear only a partial or indirect relationship to the lived reality of the individual. Therefore, in this context, suspicion attaches to a data construct, a profile that is updated and reinterpreted on a continuous basis. This reconfiguration of the legal subject also resonates with data-protection debates on profiling and automated decision-making. Once a person is translated into a score, a risk profile or a machine-readable behavioural pattern, the classificatory output may begin to structure subsequent institutional responses even before it is framed as a final decision in formal legal terms.[45]

Ontologically, the subject of suspicion becomes relational, since it is constituted through connections between transactions, accounts, locations and other factors. In other words, an individual may be rendered suspicious not because of their actions, but because of how their data correlates with patterns identified elsewhere. This is an association, not direct attribution. Most importantly, individuals may be subject to enhanced monitoring, transaction delays, account restrictions or even de-risking,[46] without a clear understanding of the justification of the measures. In such systems, the grounds for suspicion are neither transparent nor directly accessible.

Moreover, confidentiality obligations under the AML/CFT regime limit the disclosure of suspicious activity reporting.[47] Individuals are rarely informed that they have been classified as suspicious, let alone provided with meaningful reasons. Therefore, their ability to contest such classifications is constrained, since there is often no identifiable decision to challenge and no clear evidence available to dispute. AI-driven financial surveillance aggravates this situation. The case law of the Court of Justice of the European Union (CJEU) is important here because it suggests that automated scoring cannot always be treated as legally irrelevant until the moment of formal refusal or restriction.[48] Where a score substantially shapes the decision-making environment, the distinction between preliminary classification and a legally significant decision becomes much harder to sustain.[49]

The datafied subject may be left defenceless against asymmetric visibility. On the one hand, the subject and their financial activities are rendered increasingly transparent to institutions. On the other hand, the processes of evaluation remain opaque. This asymmetry is structural because visibility is a condition of regulation, while opacity is a feature of the systems that enable supervision.[50] These dynamics raise questions about agency and identity. If suspicion is generated through AI-driven financial surveillance beyond the individual's knowledge or control, the capacity to manage one's legal and social identity becomes constrained. In other words, individuals cannot anticipate how their actions will be interpreted within complex AI systems, nor can they affect the data on which those interpretations will be based. The risk is greater where the data reflect existing inequalities, institutional biases or geopolitical issues. For example, some regions, sectors or demographic groups may be disproportionately represented in risk models, amplifying broader patterns of financial inclusion and exclusion.[51] Therefore, the mechanisms through which individuals can seek accountability or redress become increasingly important, as we will see in the final section.

---

[44] Barassi, "Datafied Citizens."
[45] Aza, "Scores as Decisions?"; Case C-634/21, *SCHUFA Holding*.
[46] Satovich, "Unintended Consequences."
[47] Dabor, "Examination of the UK SAR Regime," 107.
[48] Case C-634/21, *SCHUFA Holding.*
[49] Aza, "Scores as Decisions?".
[50] Barassi, "Datafied Citizens"; Aza, "Scores as Decisions?"; Case C-634/21, *SCHUFA Holding*.
[51] Caplan, "Financial Exclusion."

## 6. Accountability Without Borders? Legal and Normative Tensions

How can accountability be articulated in situations where both the production of suspicion and the exercise of regulatory power are distributed across data infrastructures and jurisdictions? Traditionally, accountability is anchored in identifiable relationships: between a decision-maker and a subject, as well as between an act and its legal basis. Mechanisms of review, contestation and redress focus on these relationships. This is also the case with complex regulatory regimes such as AML/CFT, where multiple actors are involved. [52]

AI-driven financial surveillance complicates these assumptions. A decision affecting an individual, such as the restriction of a transaction or the closure of an account, may be the product of data processing in multiple locations, models developed by third parties and regulatory expectations originating in different legal systems. In similar cases, responsibility becomes dispersed because the locus of decision-making cannot easily be identified. This gives rise to a form of fragmented accountability. First, financial institutions remain formally responsible for compliance decisions, but their reliance on AI systems limits their ability to fully explain or justify those decisions. Second, technology providers design and maintain the AI systems, but they do not have immediate legal relationship with affected individuals. Third, regulators set expectations and standards, but they lack visibility into the operational details of the AI systems. Therefore, each actor exercises a degree of control, but no single actor possesses comprehensive oversight.

Existing legal frameworks cannot provide an effective solution to this problem. AML/CFT regimes prioritise confidentiality and prevention,[53] which translates into expansive data collection, continuous monitoring and the sharing of information across institutions and borders. For their part, data protection laws[54] offer tools such as access rights and safeguards against automated decision-making, but they are not always aligned with the specificities of financial surveillance, particularly where exemptions apply. As already mentioned, the cross-border nature of data processing introduces additional complexities because legal standards differ between jurisdictions. In the European context, this tension becomes especially visible at the intersection of AML/CFT obligations, data-protection law and the AI Act. The GDPR restricts solely automated decision-making that produces legal or similarly significant effects, while the CJEU's judgment in *SCHUFA Holding* sharpens the legal relevance of scoring and profiling by recognising that a score may itself fall within Article 22 GDPR where it plays a determining role in a third party's decision.[55] At the same time, the EU *AI Act* strengthens expectations around human oversight, documentation and accountability but, by itself, does not resolve the diffuse allocation of responsibility in cross-border surveillance infrastructures.[56]

Moreover, accountability, due process and the right to contest decisions presuppose a degree of clarity regarding how and why decisions are made. This clarity is increasingly difficult to achieve, as suspicion becomes probabilistic and relational. The practical implementation of explainability requirements also remains challenging.[57] Even where explainability is required and disclosure is permitted, the explanations provided may be partial, technical or difficult to interpret, due to the complexity of AI-driven financial surveillance. In such systems, the right to explanation risks becoming formal rather than substantive. Another difficulty lies in the translation of legal norms into machine-learning design choices. Legal requirements such as fairness, proportionality, explainability and meaningful human oversight do not map directly onto technical variables, but must be operationalised through proxies, thresholds, metrics and validation procedures. Accountability must therefore be examined not only at the level of outcomes, but also at the level of model design.[58]

These developments call for a rethinking of how accountability is understood in data-driven environments. First, greater attention must be paid to the infrastructural dimension of regulation. In other words, accountability must extend beyond individual decisions to the systems that produce them (design, data governance, etc.). Second, mechanisms of oversight must be adapted to address transnational configurations of power, potentially through the development of shared standards for algorithmic governance.[59] Third, the position of the individual within these systems requires reconsideration, particularly in relation to access to information and the ability to contest classifications in contexts where decision-making is diffuse and opaque.[60]

---

[52] Juntunen, "Accountability."
[53] Van Den Broek, "EU Preventive AML/CFT Policy."
[54] Bartlett, "Beyond Privacy," 104.
[55] More specifically, in *SCHUFA Holding*, the CJEU held that automated credit scoring by a credit reference agency may constitute "automated individual decision-making" under Article 22 GDPR where a third party relies strongly on that score in deciding whether to establish, implement or terminate a contractual relationship (e.g. bank loan). The judgment thus adopts a functional approach to algorithmic decision-making, looking beyond the formal identity of the final decision-maker to the practical influence of automated processing on decisions producing legal or similarly significant effects.
[56] Pavlidis, "EU AI Act."
[57] Preece, "Asking 'Why' in AI."
[58] Hanson, "Law-Machine Learning Translation Problem."

What is ultimately at stake is not simply whether AI enables the detection of more suspicious activity, but when and under what conditions machine-generated associations and inferences should be allowed to produce legal or regulatory consequences across borders.

## 7. Concluding Remarks: Governing Suspicion Without Borders

This article has examined how the development of AI-driven financial surveillance is transforming the nature of suspicion, the reach of regulatory authority and the constitution of the legal subject. It is argued that suspicion can no longer be understood merely as a legal judgement, tied to identifiable acts within defined jurisdictions. Instead, suspicion is increasingly produced through data-driven infrastructures that operate across borders, rendering individuals visible and governable within the continuous space of financial surveillance. Regarding the regulatory reach, this article has developed the concept of algorithmic extraterritoriality, which captures how regulatory power is increasingly exercised through the capacity to access and process cross-border data rather than through formal assertions of jurisdiction. In other words, what becomes visible within data infrastructures becomes subject to governance, regardless of territorial location. Regarding the human subject of regulation, the use of data profiles and algorithmic inferences creates datafied subjects of suspicion. Suspicion attaches to patterns and associations rather than specific acts, while it is recalibrated as new data become available. The key element is exposure to opaque processes of classification and risk assessment. These transformations place strain on existing frameworks of accountability; indeed, AI-driven financial surveillance complicates efforts to identify decision-makers, to provide meaningful explanations and to ensure effective remedies. While tools such as data protection, transparency and supervisory cooperation remain important, they cannot capture the infrastructural and transnational nature of AI-driven financial surveillance. Therefore, a rethinking is required, recognising the role of data infrastructures in shaping regulatory power. This does not imply the abandonment of established legal principles, but rather their adaptation. More specifically, if suspicion is increasingly produced through distributed systems, accountability must extend to the design, operation and governance of these systems.[61] If individuals are treated as datafied subjects, legal frameworks must ensure meaningful avenues for understanding and contesting the classifications that affect them.[62] Evidently, law must continue to articulate limits and protections in a world where the production of suspicion and the exercise of power become part of AI infrastructures, which affect contemporary social and economic life. To understand how legal concepts must evolve in response to technological change, suspicion should be reframed as an infrastructural and transnational phenomenon, while regulatory reach should be considered through the lens of algorithmic extraterritoriality.

**Acknowledgement**
This research was funded by the European Union. Views and opinions expressed are, however, those of the authors only and do not necessarily reflect those of the European Union or the European Education and Culture Executive Agency (EACEA). Neither the European Union nor EACEA can be held responsible for them.

**AI Disclosure Statement**
The authors used OpenAI's ChatGPT during manuscript preparation for language editing, stylistic refinement, and improving the clarity and structure of selected passages. The authors maintained full oversight of its use and independently reviewed, verified, and edited all AI-assisted output. AI tools were not used to generate or analyse research data or to make substantive research decisions. The authors take full responsibility for the content of the manuscript.

[59] Villarino, "Global Standard-Setting for AI," 180.
[60] Vaassen, "AI, Opacity," 2.
[61] Pavlidis, "From AI Act," 8.
[62] Pavlidis, "EU AI Act," 126.